\documentclass{article}
\usepackage{ijcai24}

\usepackage{makecell}
\usepackage{times}
\usepackage{soul}
\usepackage{url}
\usepackage[hidelinks]{hyperref}
\usepackage[utf8]{inputenc}
\usepackage[small]{caption}
\usepackage{graphicx}
\usepackage{amsmath}
\usepackage{amsthm}
\usepackage{booktabs}
\usepackage{algorithm}
\usepackage{algorithmic}
\usepackage[switch]{lineno}
\usepackage{algorithm}
\usepackage{algorithmic}
\usepackage{multirow}
\usepackage{booktabs}
\usepackage{graphicx}
\usepackage{diagbox}
\usepackage{amsmath}
\usepackage{xcolor}
\usepackage{notoccite}
\usepackage{amssymb}

\title{Ranking-based Adaptive Query Generation for DETRs in Crowded Pedestrian Detection}

\author{
    Feng Gao, Jiaxu Leng, Ji Gan, Xinbo Gao
    \affiliations
    Chongqing University of Posts and Telecommunications
    \emails
    d210201005@stu.cqupt.edu.cn, {lengjx, ganji, gaoxb}@cqupt.edu.cn
}

\begin{document}

\maketitle
\begin{abstract}
The variants of DEtection TRansformer (DETRs) have achieved promising performance in crowded pedestrian detection.
However, in crowded pedestrian detection, the performance of DETRs is sensitive to a hyper-parameter, the number of queries, which hinders its application in reality.
To settle the issue, we propose the Ranking-based Adaptive Query Generation method (RAQG), which consists of three parts, a ranking prediction head, a query supplementer, and the Soft Gradient L1 Loss (SGL1).
The core idea of RAQG is that utilizing image features to predict the number of queries replaces the manual adjustment.
Specifically, the ranking prediction head utilizes the image features produced by the transformer encoder to predict the minimum number of queries. 
Only using the minimum number of queries can lead to the imbalance of positive and negative training samples and result in missing detections.
Therefore, we design a query supplementer to improve the issue via supplementing queries.
Finally, we propose a loss function called Soft Gradient L1 Loss (SGL1) for training the ranking prediction head.
Our method is simple and light, which can be plugged into any DETRs for crowded pedestrian detection in theory. 
The experimental results on the Crowdhuman dataset and the Citypersons dataset show that our method can adaptively generate queries for DETRs and achieve competitive results. Especially, our method achieves state-of-the-art 39.4\% MR on the Crowdhuman dataset.
\end{abstract}

\section{Introduction}

\begin{figure*}[!t]
\centering
\includegraphics[width=7in]{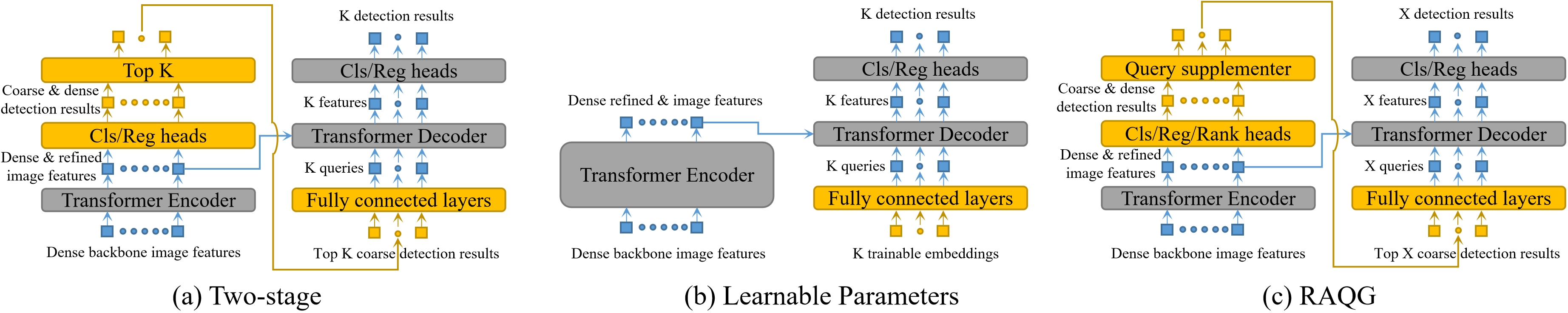}
\caption{Framework comparison of the different query generation methods. The icons with \textcolor{brown}{brown} background are the parts of the query generation methods. Since the backbone is unrelated to query generation, we omitted it in the figure.
(a) 
Two-stage treats the transformer encoder as a region proposal network to produce coarse and dense detection results. Subsequently, the K highest score detection results are selected for generating K queries.
(b) Learnable Parameters utilizes K trainable embeddings to generate K queries.
(c) Our RAQG first predicts the number of queries, X, and selects X highest score detection results to generate X queries. The K is a hyper-parameter in (a)(b) that needs to be adjusted manually. The X is predicted in (c).
\label{tab:table}}
\label{fig1}
\end{figure*}
Pedestrian detection is a significant research direction of object detection, which aims at finding where pedestrians are in an image. It is widely applied in many fields, such as video surveillance, autonomous driving, and robotics. Its downstream tasks include object tracking \cite{motr}, person re-identification \cite{reid}, gait recognition \cite{gait}, and so forth.

Currently, the performance of pedestrian detection in conventional scenes is approaching saturation. For example, the CNN-based pedestrian detector, VLPD \cite{VLPD}, has achieved 9.4\% MR on the Reasonable (R) subset of the Citypersons dataset \cite{Citypersons}. However, the performance of VLPD on the crowded Heavy Occlusion (HO) subset of the Citypersons dataset is severely limited whose MR is only 43.1\%. The reason is that most CNN-based pedestrian detectors adopt the one-to-many label assignment strategy, so a pedestrian usually corresponds to multiple detection results. The redundant detection results need to be suppressed by non-maximum suppression (NMS) based on the degree of overlaps. In crowded pedestrian detection, due to heavy overlaps among pedestrians, NMS inevitably suppresses the detection results which should be retained.

Recently, an NMS-free object detector, DETR \cite{DETR}, was proposed, which adopted the one-to-one label assignment strategy. Subsequently, DETRs \cite{DDQ,selecting} achieved state-of-the-art performance in crowded pedestrian detection. Although the issue of NMS has been resolved, new issues have emerged.
One of them is that DETRs are sensitive to the specific hyper-parameter, the number of queries, in crowded pedestrian detection.
For example, on the Citypersons dataset and Crowdhuman dataset \cite{Crowdhuman}, the number of queries must be set to 500 and 1000, respectively, otherwise, the performance is inferior to the CNN-based baseline, FPN \cite{FPN}.


As shown in Figure \ref{fig1}, the queries are one of the inputs of the transformer decoder, which are generated by the query generation method. 
The function of queries is to extract dense and refined image features in the transformer decoder. 
Then, the transformer decoder outputs K features to the classification and regression heads.
Finally, the heads output K detection results. 
Currently, there are two methods to generate queries the number of which is manually adjusted. 
The one is Two-stage which selects the K highest score coarse and dense detection results (predicted bounding boxes) $b\in \mathbb{R}^{4\times K}$ of the transformer encoder to generate K queries by fully connected layers, as shown in Figure \ref{fig1}(a) \cite{DDETR}.
The other is Learnable Parameter which uses K trainable embeddings $e\in \mathbb{R}^{E\times K}$ to generate K queries by fully connected layers, as shown in Figure \ref{fig1}(b) \cite{DETR}.
\textbf{Like Two-stage, we also use the detection results of the transformer encoder to generate queries, but the number of queries in our RAQG is predicted instead of manually adjusted, as shown in Figure \ref{fig1}(c).}

In this paper, we first conducted experiments on Learnable Parameters and Two-stage. The main purpose is to intuitively demonstrate the impact of the number of queries for DETRs in crowded pedestrian detection and the necessity of our research. Secondly, we want to show the entire process of RAQG from thinking to design. And then, we propose RAQG. Specifically, we design a ranking prediction head that can predict the ranking of the lowest score positive training sample produced by the transformer encoder. The ranking represents the minimum number of queries required to detect all pedestrians. Then, we design a query supplementer that is used to address the imbalance of positive and negative training samples and missing detection caused by the minimum number of queries. For training the ranking prediction head, we design SGL1. Our method is simple, and light, which can be plugged into any DETRs for crowded pedestrian detection in theory. 
The experimental results on Crowdhuman and Citypersons show that our method can adaptively generate queries for DETRs and achieve competitive results. Especially, our method achieves state-of-the-art 39.4\% MR on Crowdhuman. The main contributions of this work can be summarized as:
\begin{itemize}
\item Revealing the reasons why the number of queries affects performance in crowded pedestrian detection and summarizing guidelines to guide our design.

\item Proposing RAQG, which eliminates the need for manual adjustment of the number of queries in DETRs for crowded pedestrian detection.

\item Achieving state-of-the-art performance on Crowdhuman dataset with introducing negligible overhead.

\end{itemize}

\section{Related Work}

In the past decade, CNN-based detectors have dominated crowded pedestrian detection. The researches can be divided into loss functions \cite{autopedestrian,repulsion}, feature enhancement \cite{vpd}, and improving NMS \cite{N2NMS,MIP}. 
Due to the limitation of NMS, the progress is becoming increasingly slow.

The recently proposed NMS-free DETR \cite{DETR} eliminates the limitation.
Subsequently, the pioneer \cite{PETR} introduced DETRs into crowded pedestrian detection. 
Although DETRs do not require NMS for post-processing, they are prone to producing false positives.
\cite{IDDETR} and  \cite{DDQ} both argued that queries extract similar features in crowded scenes, which leads to false positives. Therefore, \cite{IDDETR} proposed a relational network to solve the problem, which utilizes contextual information to optimize the queries.  \cite{DDQ} proposed using multiple high-threshold NMS for suppressing similar queries to optimize the problem. \cite{selecting} argued that the original label assignment algorithm of DETRs assigns unlearnable labels to queries in crowded scenes, which leads to false positives. Therefore, they proposed a constraint-guided label assignment algorithm.

From our perspective, the problem is related to the number of queries. The relevant analysis is presented in section 3.2. RAQG not only achieves adaptive query generation but also optimizes the problem of false positives.


\section{Analysis of existing query generation methods in crowded pedestrian detection}
In this section, we conduct experiments on Learnable Parameters and Two-stage. The main purpose is to intuitively demonstrate the impact of the number of queries for DETRs in crowded pedestrian detection and the necessity of our research. Secondly, we want to show the entire process of RAQG from thinking to design.

\begin{figure}[!t]
\centering
\includegraphics[width=3.3in]{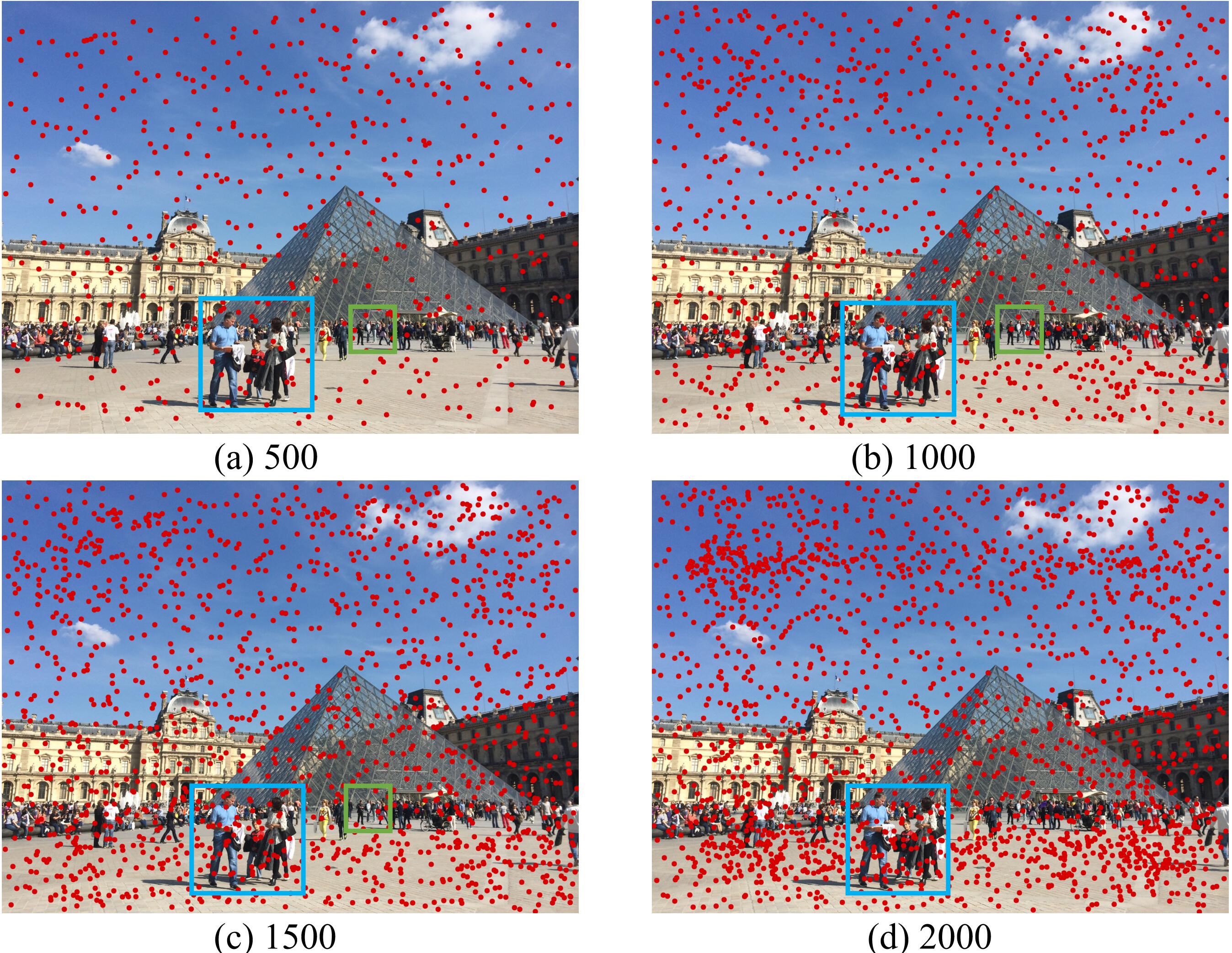}
\caption{The visualization of the different number of queries generated by Learnable Parameters. \label{tab:fig2}}
\label{fig2}
\end{figure}

\subsection{Pipeline of DETRs}
First, we introduce DETRs' pipeline to help readers understand our research. We utilize Figure 1 for illustration.
The pipeline can be formulated as:
\begin{equation}
\begin{aligned}
   x^{bac}& \gets Backbone(img)  \\
   x^{enc},(detection\thinspace results) & \gets Encoder(x^{bac}) \\
   queries & \gets query\thinspace generation\\
x^{dec} & \gets Decoder(x^{ecn},queries) \\
detection\thinspace results & \gets Head(x^{dec}) \\
\end{aligned}
\end{equation}
First, an image $img\in \mathbb{R}^{3\times H \times W}$ is input into the backbone to output the dense backbone image features $x^{bac}\in \mathbb{R}^{C\times H_b \times W_b}$. Then, the dense backbone image features $x^{bac}$ will be flattened and input into the transformer encoder to refine. Then, the transformer encoder will output the dense and refined image features $x^{enc}\in \mathbb{R}^{C\times H_bW_b}$. If the query generation method is Two-stage, besides $x^{enc}$, the transformer encoder will output $H_bW_b$ coarse and dense detection results for query generation. Then, the query generation method generates queries for extracting the dense and refined image features $x^{enc}$ in the transformer decoder. 
The forms of queries are various. However, for crowded pedestrian detection, most work adopt Deformable DETRs \cite{DDETR} as a baseline, so the form of queries is anchor boxes $r\in \mathbb{R}^{4\times K}$ and the corresponding embeddings $e\in \mathbb{R}^{E\times K}$.
Finally, the transformer decoder outputs features $x^{dec}\in \mathbb{R}^{C\times K}$, and these features $x^{dec}$ are input into the classification and regression heads to produce detection results. 
Our work focuses on query generation in the pipeline. The goal is to eliminate the hyper-parameter, the number of queries, in query generation.


\subsection{Analysis of Learnable Parameters}
We apply Learnable Parameters to Iter-Deformable-DETR \cite{IDDETR} and set the number of queries as 500, 1000, 1500, and 2000, respectively. The results are shown in Table 1. The most important evaluation metric, MR, significantly degenerates when the number of queries is 500, which is seriously inferior to FPN. There is not an obvious difference between 1000 and 1500. There is a slight degeneration when the number is 2000. 
This experiment demonstrates that the number of queries affects the performance in crowded pedestrian detection.

To better analyze how the number of queries affects the performance, we visualize these queries, as shown in Figure 2. For the sake of presentation, we simplify the queries to the center points of the anchor boxes. Firstly, these queries are uniformly and sparsely distributed on the image, which leads to a huge number of queries in the background area. Excessive queries in the background area are unnecessary. Secondly, some pedestrians are not covered by queries such as the pedestrians in the green box. Without queries, features of pedestrians are difficult to extract so pedestrians are prone to missed detection. Thirdly, as shown in the blue box in Figure 2(d), large-scale pedestrians are likely to be covered by multiple queries. These dense queries are likely to extract similar features, but the labels for these queries are different due to the one-to-one label assignment. In the training process of object detection, it needs a label assignment algorithm to assign a label to each detection result. For DETRs, detection results come from queries. The queries that extract similar features are likely to produce similar detection results. The one-to-one label assignment indicates that one pedestrian just matches one detection result. The rest detection results match the background. The queries that extract similar features match different labels, which can easily lead to false positives. In addition, detection results matched with pedestrians are called positive training samples. Detection results matched with the background are called negative training samples. The imbalanced proportion of positive and negative training samples can not be ignored. As shown in Figure 3(a), there are 7 persons. If the number of queries is 1000. There will be 7 positive and 993 negative training samples, which is imbalanced.

\begin{figure}[!t]
\centering
\includegraphics[width=3.3in]{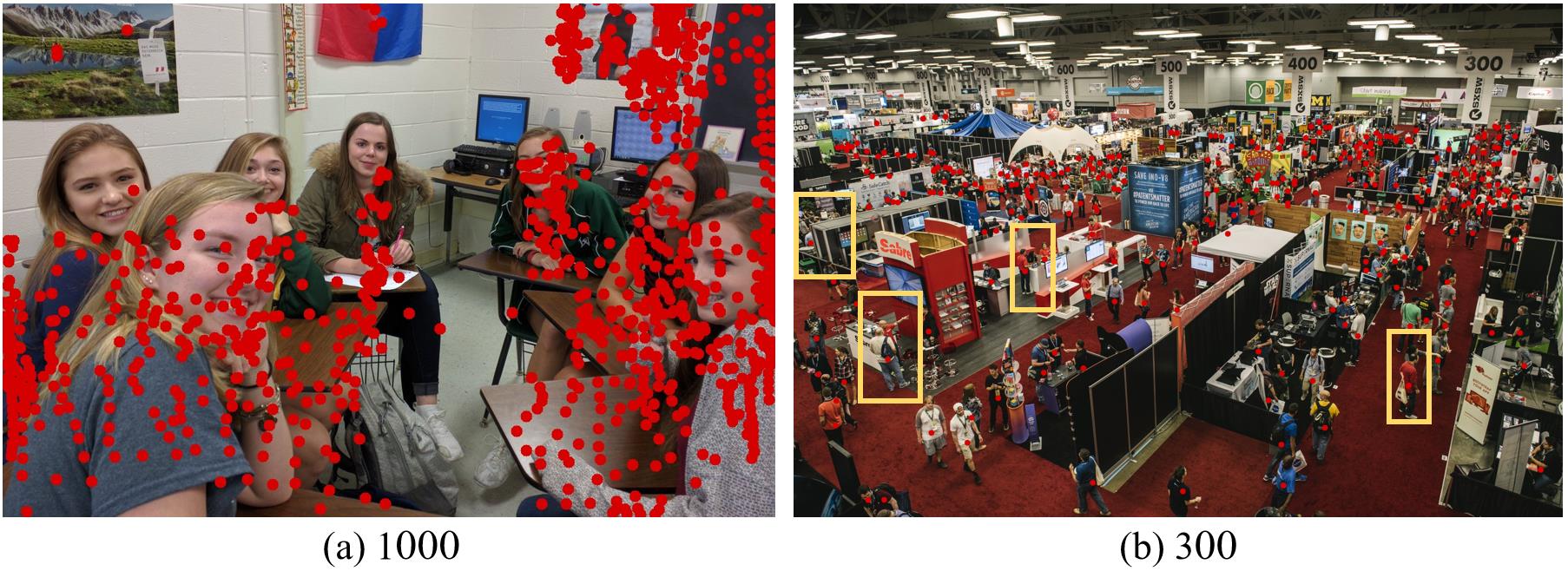}
\caption{The visualization of the different number of queries generated by Two-stage. \label{tab:table}}
\label{fig2}
\end{figure}

\begin{table}[!t]
\caption{The performance on the Crowdhuman dataset. \label{tab:table}}
\centering
\resizebox{\linewidth}{!}
{
\begin{tabular}{ccccc}
\toprule[1.4pt]
Method&Queries &AP$\uparrow$ &MR$\downarrow$&Recall$\uparrow$ \\
\hline
FPN \cite{FPN}&- &85.7& 42.9 &-\\
\hline
\multirow{4}{*}{Learnable Parameters}&500 &87.1& 52.0&94.0 \\

&1000 &92.0& 41.5 &96.4\\

&\textcolor{red}{1500} &\textcolor{red}{92.2} &\textcolor{red}{41.4}&\textcolor{red}{96.6}\\

&2000 &92.2 &42.6&96.6\\

\hline
\multirow{3}{*}{Two-stage}&\textcolor{red}{300} &\textcolor{red}{92.1}& \textcolor{red}{41.7}&96.6 \\

&500 &92.0& 43.6 &97.0\\

&1000 &91.7 &45.0&\textcolor{red}{97.5}\\

\toprule[1.4pt]
\end{tabular}}
\end{table}

\begin{figure*}[t]
\centering
\includegraphics[width=7in]{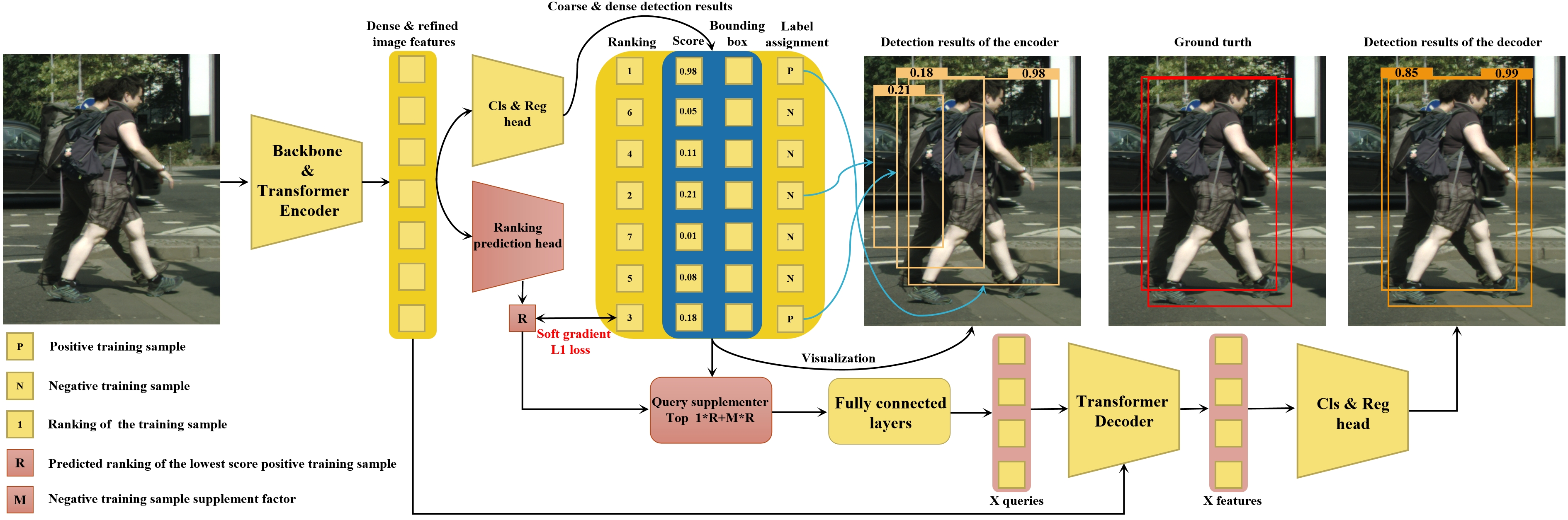}
\caption{The pipeline of DETRs equipped with our RAQG. The part with \textcolor{red}{red} background is RAQG. The part with \textcolor{yellow}{yellow} or \textcolor{blue}{blue} background is the common part of DETRs. \label{tab:table}}
\label{fig2}
\end{figure*}

\subsection{Analysis of Two-stage}
We apply Two-stage to Iter-Deformable-DETR. When the number of queries is 500 or 1000, the performance of MR is also inferior to FPN. 
We visualize the queries of Two-stage, as shown in Figure 3. In Figure 3(a), excessive queries are concentrated on pedestrians, which can lead to false positives. Although reducing the number of queries contributes to alleviating false positives, when the pedestrians are more than the queries, the excess pedestrians cannot be missed in detection. As shown in Figure 3(b), the number of pedestrians exceeds the number of queries.
The pedestrians in the yellow boxes will be missed in detection due to a lack of queries.
Additionally, in Figure 3(b), the positive and negative training samples are also imbalanced because there is no negative training sample.

\begin{table}[!t]
\caption{The performance on the Citypersons dataset.}
\label{table3}
\centering
\begin{tabular}{cccc}
\toprule[1.4pt]
Method&Queries &R$\downarrow$ &HO$\downarrow$\\
\hline
\multirow{4}{*}{Learnable Parameters}

&200 &12.6& 43.1 \\

&300 &11.1& 42.6 \\

&\textcolor{red}{500} &\textcolor{red}{10.4}& \textcolor{red}{40.8} \\

&700 &11.6& 43.2 \\


\hline
\multirow{3}{*}{Two-stage}

&100 &10.8& 41.6 \\

&\textcolor{red}{200} &\textcolor{red}{10.8}& \textcolor{red}{41.0} \\

&300 &11.2& 42.5 \\

\toprule[1.4pt]
\end{tabular}
\end{table}

\subsection{Summary}
The same experiment was conducted on the Citypersons dataset. The best number of queries is 500 for Learnable Parameters and 200 for Two-stage, as shown in Table 2. 
The results of Tables 1 and 2 demonstrate that it is necessary to propose an adaptive query generation method to eliminate the impact of the number of queries for DETRs in crowded scenes. Based on the above analysis, we summarize four guidelines for designing adaptive query generation methods:

\begin{itemize}
\item The number of queries should not be less than the number of pedestrians.

\item The queries covered on the same pedestrian should not be too dense to avoid extracting highly similar features.

\item A pedestrian needs a query nearby at least for feature extraction.

\item The proportion of queries for pedestrians and backgrounds should be appropriate.

\end{itemize}

\section{Ranking-based Adaptive Query Generation}


\subsection{Ranking prediction head}
Based on the four guidelines, we adopt a divide-and-conquer strategy to design RAQG. The first step is to only follow the first three guidelines.

In the training process of object detection models, there is a step called label assignment mentioned in section 3.2, which uses an algorithm to assign a label to each detection result. As shown in Figure 4, the blue lines show the process. 
For simplicity, we only present the three training samples with the highest scores.
We find that using all samples with a score not less than the lowest positive training sample to generate queries can satisfy the first three guidelines.
Taking Figure 4 as an example, the lowest score of the positive training samples is 0.18. The corresponding ranking is 3. We utilize the ranking 3 as the number of queries and select the 3 highest score detection results to generate 3 queries. The number of queries, 3, is not less than the number of pedestrians, 2, which satisfies the first guideline. The sparse distribution of three queries satisfies the second guideline. The two queries generated by the two positive training samples have high overlaps with the two pedestrians, which satisfies the third guideline.

Therefore, we designed a tiny network called ranking prediction head to predict the ranking of the lowest score positive training sample. Then, based on the predicted ranking R, we select the R samples with the highest scores to generate the queries adaptively.

The network structure of our rank prediction head is shown in Figure 5, which contains a multi-head self-attention module (MHSA) \cite{Attention}, three fully connected layers, and a Relu activation function. The function of MHSA is to pool the dense and refined image features $x^{enc}\in \mathbb{R}^{C\times H_bW_b}$ into a feature $x^{mhsa}\in \mathbb{R}^{C\times 1}$. MHSA contains 3 inputs, Q, K, and V. The Q is the mean of the dense and refined image features $x^{enc}_{m}\in \mathbb{R}^{C\times 1}$ and the K and the V are the dense and refined image features $x^{enc}$. The three fully connected layers mimic the network structure of DETRs' regression head. 
The number of input and output channels for the first two fully connected layers is C and for the last fully connected layer is C and 1.
The function of Relu is to constrain the range of R between 0 and positive infinity due to the ranking can not be minus.

\subsection{Query supplementer}
Directly using R queries will result in almost no negative training samples in the transformer decoder, as the score of positive training samples is usually higher than that of negative training samples, which does not satisfy the fourth guideline. Therefore, we design a query supplementer to supplement negative training samples for the transformer decoder. The query supplementer is based on our conjecture that positive and negative training samples should be an appropriate proportion and the appropriate proportion is a constant. 
Therefore, we select the 1$\times$R+M$\times$R highest score detection results to generate queries for the transformer decoder. The M is the proportion of positive and negative training samples.

In section 5.2, our experiments demonstrate that the optimal value of M is 5. Furthermore, using 1+M times the ranking as the label for R to train the ranking prediction head can remove the query supplementer and achieve the same performance.

\subsection{Generating queries}
The process of converting the selected detection results into queries is the same as that of Two-stage.
First, we select the X highest score detection results (predicted bounding boxes) $b\in \mathbb{R}^{4\times X}$ based on the prediction of the ranking prediction head. Then, $b\in \mathbb{R}^{4\times X}$ is as the anchor boxes of the queries. Finally, we input $b\in \mathbb{R}^{4\times X}$ to the fully connected layers to generate the corresponding embeddings $e\in \mathbb{R}^{E\times X}$. 

\subsection{Soft Gradient L1 loss}
The rank prediction head and the regression head are both used for regression tasks in DETRs. However, their regression ranges are quite different. 
The regression range of the regression head is from 0 to 1 and the regression range of the ranking prediction head is from 0 to positive infinity.
Utilizing the same L1 loss to train the two heads is unstable. 
Deep learning trains models through backpropagation.
In the process of backpropagation, it is the gradient to update the model parameters. 
The process of backpropagation can be formulated as
\begin{equation}
\begin{aligned}
   w_{t}& \gets w_{t-1}-\eta \cdot g_{w} \cdot g_{l}, 
\end{aligned}
\end{equation}
where $w$ is the parameters of the model and $t$ is the training step and $\eta$ is the learning rate and $g_{l}$ is the gradient produced by the loss function and $g_{w}$ is the rest gradients. 
L1 Loss only contains three discrete gradients as follows: 
\begin{equation}
g_{l}=\left\{\begin{matrix} 

  1, y^{*}-y>0 \\
0, y^{*}-y=0  \\
  -1,y^{*}-y<0 ,  \\
\end{matrix}\right. 
\end{equation}
where $y^{*}$ is the prediction and $y$ is the label. 
For example, when $y^{*}-y=1000$, it is obvious that the prediction of the ranking prediction head is inaccurate, so the parameters of the model need a severe update. According to Equation (2), the parameters of the model are updated from $w_{t} \gets w_{t-1}$. 
The update is $\eta \cdot g_{w} \cdot 1$. When $y^{*}-y=1$, for our ranking prediction task, the prediction is accurate, so the parameters of the model need to be updated hardly. However, the update is still $\eta \cdot g_{w} \cdot 1$. The discrete gradient of L1 loss is detrimental to the training of ranking prediction heads

Therefore, we propose SGL1. \textbf{Our core idea is to convert the discrete gradient $g_{l}\in [1, 0, -1]$ of L1 loss into the continuous gradient.} SGL1can be formulated as
\begin{equation}
g_{l}=\left\{\begin{matrix} 

  sigmoid(1)-sigmoid(y/y^{*}), y^{*}-y\geq0 \\
  sigmoid(y^{*}/y)-sigmoid(1),y^{*}-y<0 . \\
\end{matrix}\right. 
\end{equation}
The closer the loss value $|y^{*}-y|$ is to 0, the closer the gradient $g_{l}$ is to 0.
On the contrary, the greater the difference between the loss value $|y^{*}-y|$ and 0, the greater the difference between the gradient $g_{l}$ and 0.

With our method, the loss functions of DETRs can be formulated as
\begin{equation}
Loss=\lambda _{1}L_{cls}+\lambda_{2}L_{GIoU}+\lambda_{3}L_{L1}+\lambda_{4}L_{SGL1},
\end{equation}
where $\lambda _{1}$, $\lambda _{2}$, $\lambda _{3}$ and $\lambda_{4}$ are 2, 2, 5 and 0.05.

\begin{figure}[!t]
\centering
\includegraphics[width=3.3in]{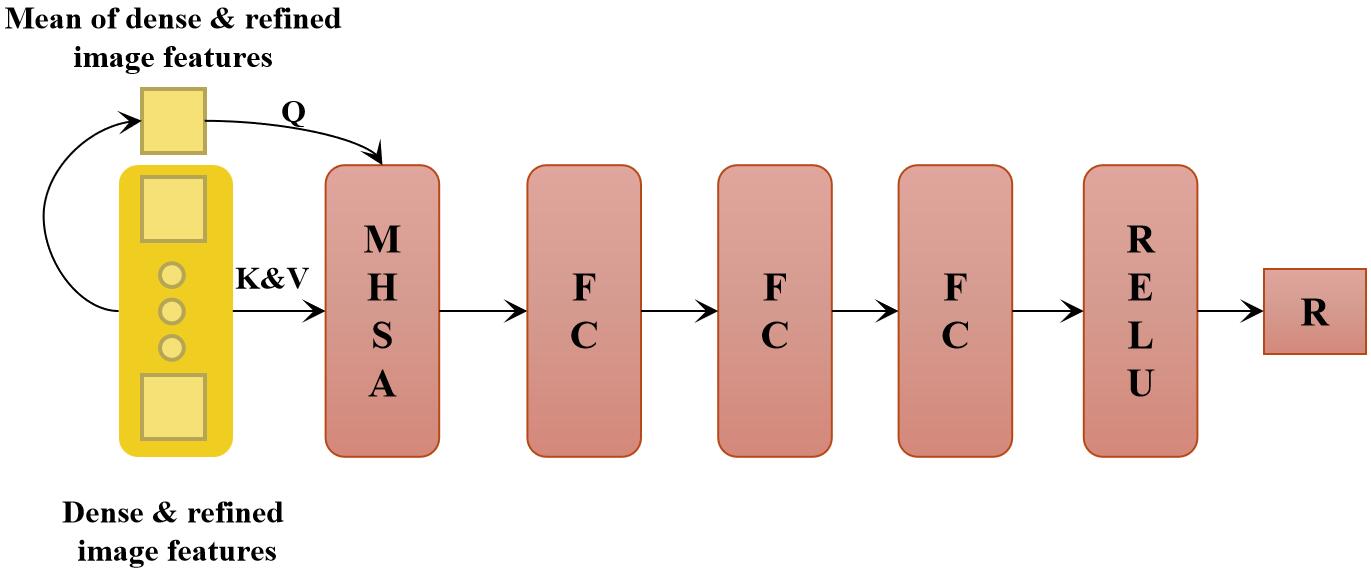}
\caption{The network of our ranking prediction head. \label{tab:table}}
\label{fig2}
\end{figure}

\section{Experiment}
\subsection{Dataset, metrics, and implementation details}
\textbf{Dataset:} We show the effectiveness of RAQG on the challenging Crowdhuman dataset \cite{Crowdhuman} and the comprehensive Citypersons dataset \cite{Citypersons}. 

\textbf{Metrics:} We follow the evaluation metrics used in Citypersons and Crowdhuman, which contain MR, AP, and Recall:

\begin{itemize}
\item Log-average Miss Rate (MR) is a commonly used and the most important evaluation metric in pedestrian detection, focusing on the number of false positives per image (FPPI) \cite{Caltech}.
\item Average Precision (AP) is a commonly used evaluation metric in general object detection, which summarizes the PR curve of the detection results. 
\item Recall is a common evaluation metric for calculating the ratio between true positive and ground truths.
\end{itemize}

\begin{table*}[!t]
\caption{Results of different M on Crowdhuman and Citypersons. The HO indicates the pedestrians over 50 pixels in height with more than 35\% occlusion. The R indicates the pedestrians over 50 pixels in height with less than 35\% occlusion. 
\label{tab:table}} 
\centering
\begin{tabular}{ccccccccccccc}
\toprule[1.4pt]
Dataset &\diagbox{Metrics}{M}&0 &1 & 2&3 &4 & 5&removal&6 &7 &8\\
\hline

\multirow{3}*{\makecell{Crowdhuman}}& MR$\downarrow$   &40.2&40.2 & 40.1&40.0 &39.8 & \textcolor{red}{39.4}& \textcolor{red}{39.4}&39.7 &39.8 & 39.9\\
                                    & AP$\uparrow$    &91.8&92.9 & 93.1&93.2 &93.2 & \textcolor{red}{93.2}& \textcolor{red}{93.4}&93.2 &93.2 & 93.2\\
                                    & Recall$\uparrow$&95.6&97.9 & 98.5&98.6 &98.6 & \textcolor{red}{98.7}& \textcolor{red}{98.7}&98.7 &98.7 & 98.7\\
\hline 
\multirow{2}*{\makecell{Citypersons}} &HO$\downarrow$&43.9 &43.1 &42.5&41.8 &41.0 & \textcolor{red}{40.4}& \textcolor{red}{40.0}&41.3&41.4 & 41.5\\
                                      &R$\downarrow$ &12.9 &11.1 &10.4&10.1 &10.0 & \textcolor{red}{9.8}& \textcolor{red}{9.8}&10.3&10.6 & 10.1\\
\toprule[1.4pt]
\end{tabular}

\end{table*}

\textbf{Implementation Details:} We test the effectiveness of RAQG on three DETRs, Deformable DETR, Iter-Deformable-DETR, and SSCP \cite{selecting}. The first one is a general object detector and the rest are crowded pedestrian detectors. We adopt ResNet-50 as their backbone. The transformer encoder and decoder are both 6 layers and 256 is the hidden dimension. For the learning rate scheduler, we use 0.0001 as an initial learning rate and drop it at the 40th epoch by multiplying 0.1 for the 60 epoch setting. We use the SDG optimizer with a weight decay of 0.0001 and train models on an RTX 3090 GPU. The batch size is 1.

\subsection{Ablation Study}
\textbf{Query Supplementer:} To verify our conjecture that M is a constant instead of a hyper-parameter that represents the appropriate proportion of positive and negative training samples. The experiment is conducted on Crowdhuman and Citypersons. We test M from 0 to 8. The consistent results are shown in Table 3. The performance is optimal on the two datasets when M is 5. 
The experiment demonstrates our conjecture that M is a constant.
Furthermore, we remove Query Supplementer and use 1+5 times the ranking of the lowest score positive training sample as the label to train the ranking prediction head. The result is shown in Table 3 with removal, which demonstrates that RAGQ can achieve adaptive query generation without any hyper-parameters.

\textbf{Soft Gradient L1 Loss:}
As shown in Table 4, Soft Gradient L1 Loss is better than the other regression losses for training the ranking prediction head. L2 Loss leads to gradient explosion due to the huge loss value. L1 Smooth is proposed in Fast-RCNN \cite{fastrcnn} for training the regression head with some hyper-parameters. For our ranking prediction head, if its hyper-parameters are not reconfigured, it is indistinguishable from L1 loss. Therefore, the results of L1 and L1 Smooth are similar.

\begin{table}[!t]
\caption{Comparison with the other regression losses. GE indicates gradient explosion. \label{tab:table}}
\centering
\resizebox{\linewidth}{!}
{
\begin{tabular}{ccccccc}
\toprule[1.4pt]
Dataset &\diagbox{Metrics}{Loss}&L1 &L2 & L1 Smooth&SGL1 \\
\hline

\multirow{3}*{\makecell{Crowdhuman}}& MR$\downarrow$    &40.0& GE&39.9 &\textcolor{red}{39.4} \\
                                    & AP$\uparrow$    &93.1& GE&93.1 &\textcolor{red}{93.4} \\
                                    & Recall$\uparrow$&98.6& GE&98.6&\textcolor{red}{98.7} \\
\hline 
\multirow{2}*{\makecell{Citypersons}} &HO$\downarrow$&41.9 & GE&41.0&\textcolor{red}{40.0}\\
                                      &R$\downarrow$ &10.1 & GE&10.0&\textcolor{red}{9.8} \\
\toprule[1.4pt]
\end{tabular}}
\end{table}

\subsection{The overhead}
We apply Learnable Parameters, Two-stage, and RAQG on SSCP to test the inference time cost on Crowdhuman. The results are 20.9, 20.4, and 20.3 frames per second (FPS).  The result demonstrates that RAQG achieves adaptive query generation with almost no additional overhead.

\subsection{The universality}
We apply RAQG on Deformable DETR, Iter-Deformable DETR, and SSCP to verify the universality of RAQG. The experimental results are shown in Table 5, which demonstrates that RAQG can adaptively generate queries on 3 DETRs and compared with the other two query generation methods, RAQG achieves better performance. 

\begin{table}[!t]
\caption{Comparison with the state-of-the-arts on Crowdhuman. $\sim$ means average. LP indicates Learnable Parameters. $\sim$ means average.\label{tab:table}}
\centering
\resizebox{\linewidth}{!}{
\begin{tabular}{ccccc}
\toprule[1.4pt]
Model-Query generation method &Queries&MR$\downarrow$&AP$\uparrow$&Recall$\uparrow$\\
\hline
Deformable DETR -LP    &1000      &44.6         &90.9         &95.5\\

Deformable DETR-Two-stage      &300      &43.1         &91.9        &96.3\\

Deformable DETR-RAQG      &$\sim$269      &\textcolor{red}{42.5}         &\textcolor{red}{92.4}            &\textcolor{red}{97.2}   \\
\hline
Iter-Deformable DETR-LP      &1000      &41.5         &92.0         &96.4\\

Iter-Deformable DETR-Two-stage      &300      &41.7         &92.1         &96.6\\

Iter-Deformable DETR-RAQG      &$\sim$266    &\textcolor{red}{41.2}         &\textcolor{red}{92.9}     &\textcolor{red}{97.5}\\

\hline
SSCP-LP      &1000      &39.7         &92.4         &97.0\\

SSCP-Two-stage      &300      &40.4         &92.4         &96.8\\

SSCP-RAQG       &$\sim$464    &\textcolor{red}{39.4}         &\textcolor{red}{93.4}     &\textcolor{red}{98.7}\\
\toprule[1.4pt]
\end{tabular}}
\end{table}

\begin{figure*}[!t]
\centering
\includegraphics[width=7in]{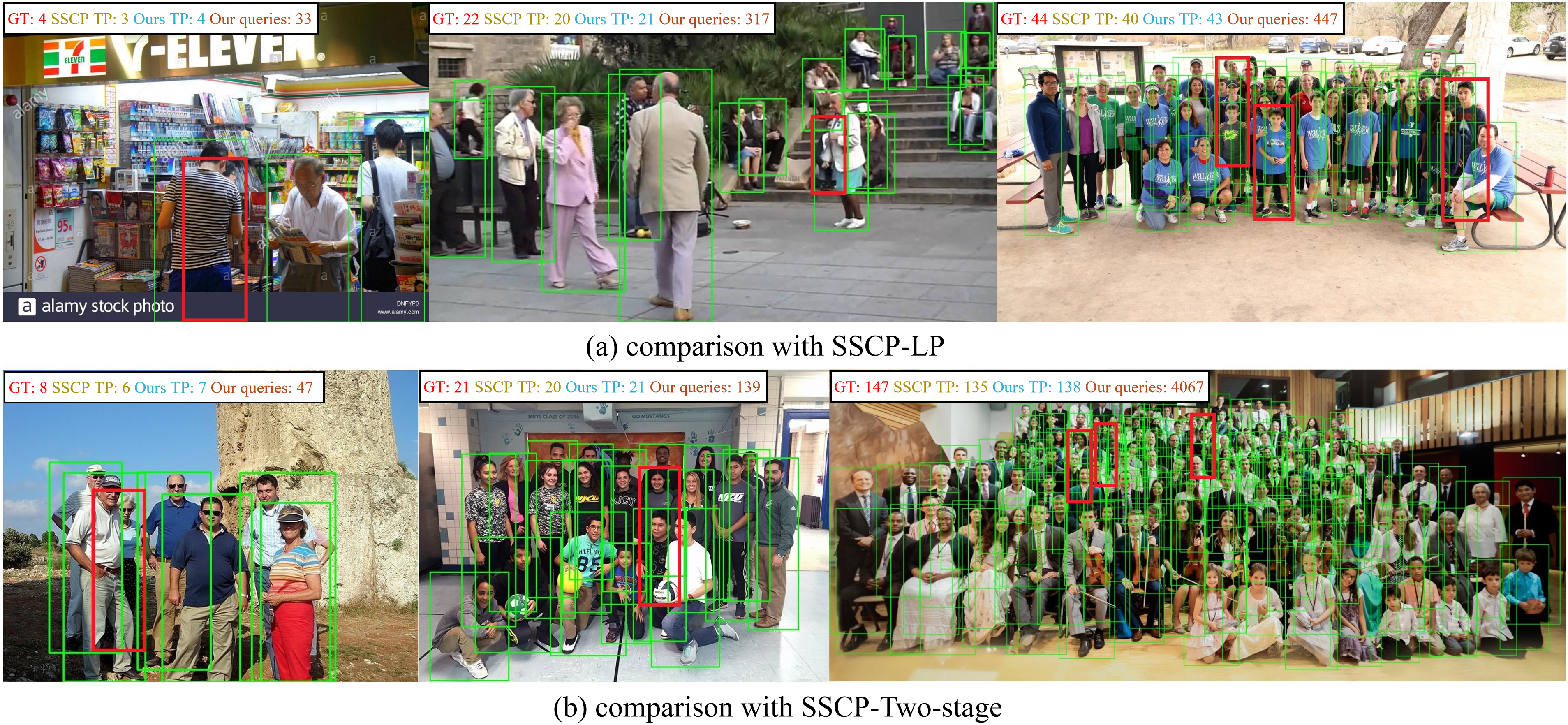}
\caption{The visualization results (at FPPI=1) of SSCP-RAQG, SSCP-LP, and SSCP-Two-stage. The green boxes are common true positives. The red boxes are true positives only produced by SSCP-RAQG.\label{tab:table}}
\label{fig2}
\end{figure*}

\subsection{Comparison with the state-of-the-art}
\textbf{Crowdhuman:} As shown in Table 6, compared with our baseline SSCP, the performance has been improved across the board. We make comparisons with the state-with-the-arts: Faster-R-CNN-FPN \cite{FPN}, CrowdDet \cite{MIP}, , AEVB \cite{vpd}, OPL \cite{song2023OPL}, OTF \cite{OTF}, and DDQ DETR \cite{DDQ}. Our method achieves state-of-the-art 39.4\% MR and 98.7\% Recall. It is worth noting that DDQ DETR achieves better AP and worse MR with the same Recall. It means that our Query-adaptive DETR produces more true positives in the high score interval which are more beneficial for downstream tasks of pedestrian detection.

\textbf{Citypersons:} We compare our methods with several state-of-the-art pedestrian detectors on Reasonable(R) subset, Heavy Occlusion(HO) subset, including FAPD \cite{FAPD}, DPFD \cite{DPFD}, FC-Net \cite{fc}, OAF-Net \cite{OAF} and VLPD \cite{VLPD}. As shown in Table 7, compared to using LP and Two-stage, SSCP using RAQG to generate queries can achieve similar performance without setting the number of queries.
Compared with the state-of-the-art methods, SSCP with RAQG has obvious advantages in severely crowded and occluded HO subset.

\begin{table}[!t]
\caption{Comparison with the state-of-the-arts on Crowdhuman. \label{tab:table}}
\centering
\resizebox{\linewidth}{!}{
\begin{tabular}{cccccc}
\toprule[1.4pt]
Model              &Backbone &Queries&MR$\downarrow$&AP$\uparrow$&Recall$\uparrow$\\
\hline
FPN (CVPR'17)      &R50      &-      &42.7         &87.1         &90.4\\

CrowdDet (CVPR'20) &R50      &-      &41.4         &90.2         &93.7\\

AEVB (CVPR'21)     &R50      &-      &40.7         &-            &-   \\




OPL (CVPR'23)      &R50      &-      &44.9         &91.0         &97.7\\

OTF (CVPR'23)      &R50      &-      &45.2         &90.9         &97.9\\



DDQ DETR(CVPR'23)  &R50      &900    &39.7         &\textcolor{red}{93.8}     &\textcolor{red}{98.7}\\

\hline
SSCP-RAQG&R50      &$\sim$464 &\textcolor{red}{39.4}&93.4&\textcolor{red}{98.7}\\
\toprule[1.4pt]
\end{tabular}}
\end{table}

\begin{table}[!t]
\caption{Comparison with the state-of-the-arts on Citypersons.\label{tab:table}}
\centering
\resizebox{\linewidth}{!}{
\begin{tabular}{cccccc}
\toprule[1.4pt]
Model            &Backbone&Queries  &R$\downarrow$ & HO$\downarrow$\\
\hline

FAPD(TITS'22)     & VGG-16&-&10.9 & 47.5\\

DPFD(TITS'22)     & VGG-16&-&10.3 & 50.2\\

FC-Net(TITS'22)   & R50&-&13.5 & 44.3\\

OAF-Net(TITS'22)  &HRNet-W32&-&9.40 & 43.1\\
VLPD(CVPR'23)     & R50    &-        &9.40          & 43.1\\
\hline 
SSCP-LP (ACM MM'23)  &R50     &500      &10.4          &40.0\\
SSCP-Two-stage   &R50     &200      &10.2          &40.5\\
SSCP-RAQG    &R50     &$\sim$83 &9.8          &40.0\\
\toprule[1.4pt]
\end{tabular}}

\end{table}

\section{Visualization}
In Figure 6, we can see the visualization of SSCP-LP, SSCP-Two-stage, and SSCP-Two-RAQG.

Figure 6(a) shows 3 images with completely different scenes, different numbers of pedestrians, and different crowded degrees. However, SSCP-LP shows the same missing detection. The missing detection tends to occur in extremely crowded areas as analyzed in Figure 2. The queries generated by Learnable Parameters are uniformly and sparsely distributed on the image. Even with 1000 queries, it can not detect all pedestrians in crowded scenes. Additionally, the right of Figure 6(a) shows that, with more pedestrians, even if the boy is not occluded by others in the middle of the image he also can not be detected, because there are no more queries around the boy. 
Excessive queries are in the background area.
By contrast, using RAQG for query generation achieves better performance with fewer queries.

Figure 6(b) is a comparison of SSCP-RAQG and SSCP-Two-stage. SSCP-Two-stage generates 300 queries and still produces missing detections. However, the reason is different. There are enough queries near pedestrians, but SSCP-Two-stage classifies them as backgrounds instead of pedestrians. when the number of pedestrians is small, selecting the top 300 queries for training will confuse the model due to one-to-one label assignment, as shown in Figure 3(a). For the right image of Figure 6(b), RAQG generates queries adaptively based on images, resulting in more true positives. The reason for the huge number of queries is that in such a crowded scene, the lowest score positive sample ranking is lower.

\section{Conclusion}
In this paper, we have analyzed the reason for the fixed number of queries impacting the performance of DETRs in crowded pedestrian detection. Based on this analysis, we proposed RAQG. RAQG contains three parts: a ranking prediction head, a query supplementer, and SGL1. To verify the effectiveness of RAQG, we plugged RAGQ into 3 DETRs and compared with multiple state-of-the-arts methods.
The experimental results show that RAQG can theoretically be plugged into any DETRs, and RAQG can not only achieve adaptive query generation without hyper-parameters and almost additional overhead but also improve performance.

\bibliographystyle{named}
\bibliography{ijcai24}

\end{document}